\documentclass[runningheads]{llncs}

\usepackage{import}
\usepackage{booktabs}           
\usepackage{multirow}           
\usepackage{amsfonts}           
\usepackage{graphicx}           
\usepackage{duckuments}         
\usepackage{tabularx}
\usepackage{adjustbox}
\usepackage{comment}
\usepackage{amsmath}
\usepackage{colortbl}
\usepackage{mathtools}
\usepackage{algorithmic}
\usepackage{blkarray, bigstrut}
\usepackage{makecell}
\usepackage[ruled,linesnumbered]{algorithm2e}
\usepackage{latexsym}
\usepackage{microtype}
\usepackage{url}
\usepackage[most]{tcolorbox}

\begin{document}

\title{Integrating Knowledge Graph Embeddings and Pre-trained Language Models\\ in Hypercomplex Spaces}

\author{Mojtaba Nayyeri\inst{1}$^*$ \and
Zihao Wang\inst{1}$^*$ \and
Mst. Mahfuja Akter\inst{2}$^*$ \and 
Mirza Mohtashim Alam \inst{3} \and
Md Rashad Al Hasan Rony \inst{2} \and 
Jens Lehmann\inst{4} \and 
Steffen Staab \inst{1,} \inst{5}
}
\institute{University of Stuttgart, Stuttgart, Germany \\
\email{\{mojtaba.nayyer, zihao.wang, steffen.staab\}@ipvs.uni-stuttgart.de} \and
University of Bonn, Bonn, Germany \\
\email{mahfuja.ruby85@gmail.com} \\
\email{s39mrony@uni-bonn.de}\and
Karlsruhe Institute of Technology, Karlsruhe, Germany\\
\email{turzo.mohtasim@gmail.com} \and 
TU Dresden, Amazon (work done outside of Amazon),
\email{jlehmnn@amazon.com} \and
University of Southampton, UK
}

\maketitle              
\def\thefootnote{*}\footnotetext{These authors contributed equally to this work.}\def\thefootnote{\arabic{footnote}}
\begin{abstract}
Knowledge graphs comprise structural and textual information to represent knowledge.
To predict new structural knowledge, current approaches learn representations using both types of information through knowledge graph embeddings and language models.
These approaches commit to a single pre-trained language model.
We hypothesize that heterogeneous language models may provide complementary information not exploited by current approaches.
To investigate this hypothesis, we propose a unified framework that integrates multiple representations of structural knowledge and textual information.
Our approach leverages hypercomplex algebra to model the interactions between (i) graph structural information and (ii) multiple text representations.
Specifically, we utilize Dihedron models with 4*D dimensional hypercomplex numbers to integrate four different representations: structural knowledge graph embeddings, word-level representations (e.g., Word2vec and FastText), sentence-level representations (using a sentence transformer), and document-level representations (using FastText or Doc2vec).
Our unified framework score the plausibility of labeled edges via Dihedron products, thus modeling pairwise interactions between the four representations.
Extensive experimental evaluations on standard benchmark datasets confirm our hypothesis showing the superiority of our two new frameworks for link prediction tasks.

\end{abstract}

\section{Introduction}
Knowledge Graphs (KGs) have become an integral part of many AI systems, ranging from question answering and named entity recognition to recommendation systems \cite{ji2021survey,choudhary2021survey,nickel2015review}.
KGs represent knowledge in the form of multi-relational directed labeled graphs, where nodes with labels can represent entities (e.g., \textit{``Q5220733''}), and labeled edges represent relations between entities (e.g., \textit{P19}).
Therefore, a fact can be represented as a triple, \textit{(node, edge label, node)}, such as \textit{(Q5220733, P19, Q621549)} in Wikidata. 

\begin{figure}[h!]
    \centering
    \includegraphics[width=10cm,height=5cm]{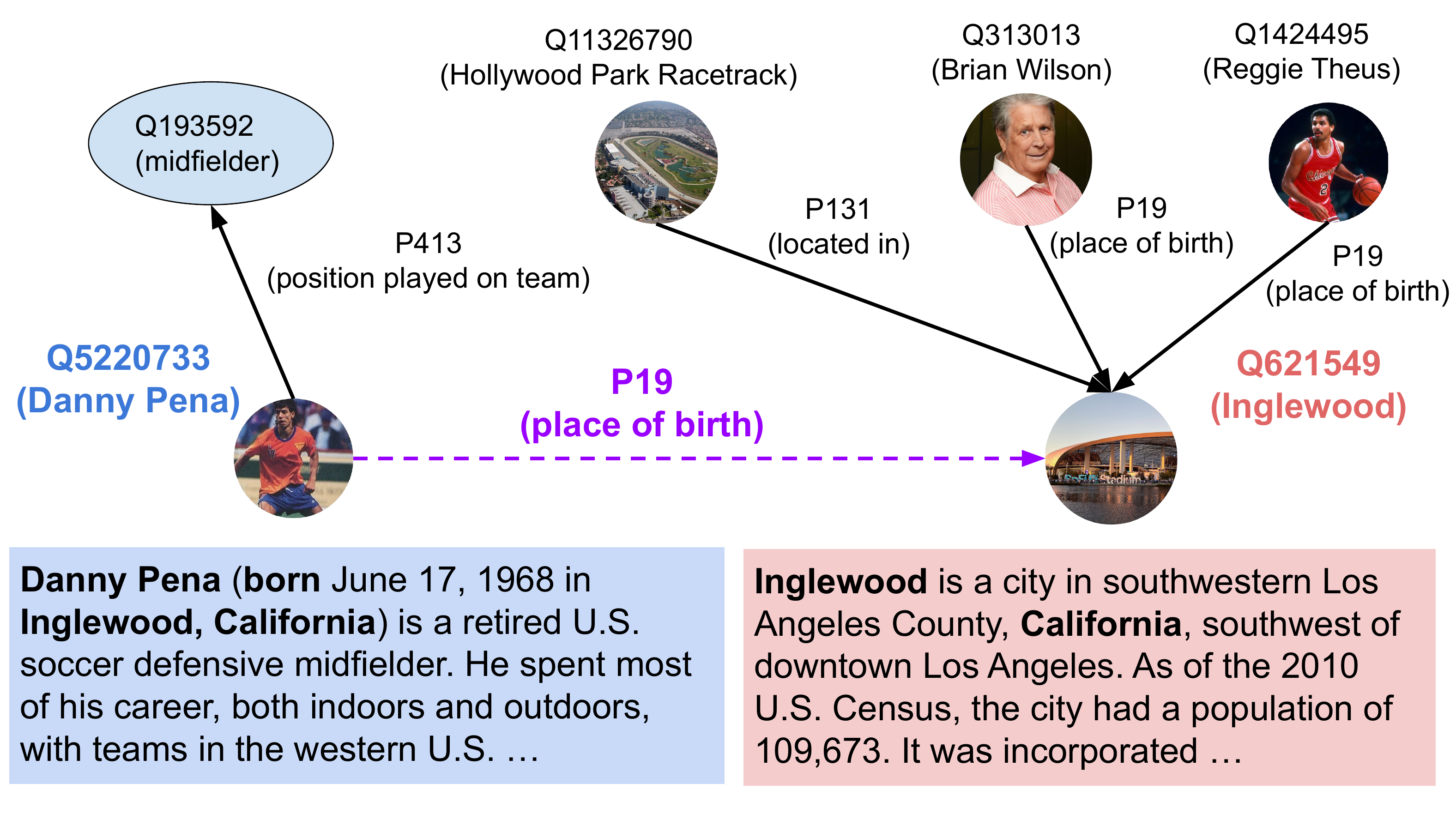}
    \caption{Knowledge Graph with textual descriptions of entities. The entity ``Q5220733'' lacks proper structural information, but it comes with a rich textual description, which may help
     to predict the place of birth.}
    \label{fig: baseMotivExamp}
\end{figure}

In order to enable machine learning to act on KGs with symbolic information \cite{wang2017knowledge}, Knowledge Graph embeddings (KGE) map nodes and edge labels to a low-dimensional vector space. These embeddings are assumed to capture semantic and structural knowledge and can support machine learning tasks such as link prediction, entity linking, and question answering.
However, despite the large number of facts contained in KGs, they are still incomplete compared to the facts that exist in the world, which can have a negative impact on downstream tasks.
Consider Figure~\ref{fig: baseMotivExamp} and assume that the purple dashed edge (Q5220733, P19, Q621549) is unknown because the entity ``Q5220733'' is only connected to one other entity, ``Q193592''.
Although structural graph information alone cannot help bridge the gap between ``Q5220733'' and ``Q621549'', a second textual representation, such as additional information from a source like Wikipedia\footnote{\url{https://en.wikipedia.org/wiki/Danny_Pena}}, could be used to provide a solution.

Early approaches such as DKRL~\cite{xie2016DKRL} and ConMask~\cite{shi2017open}, have gone beyond structural graph knowledge and incorporated textual information for link prediction using deep learning methods, such as convolutional neural networks (CNNs) and attention mechanisms to transform textual information into joint latent representations with structure-based KGEs.
More recent approaches \cite{PLMyao2019kgbert,zhang-etal-2020-pretrain,PLMli2021siamese,wang2021kepler} have incorporated pre-trained language models into KGE models by unifying their two loss functions.
However, these approaches only represent texts with a single pre-trained language model, which may lead to inferior performance when encoding textual information in certain KGs.
For example, while BERT is employed in \cite{PLMyao2019kgbert,zhang-etal-2020-pretrain,wang2021kepler}, evaluations \cite{DBLP:conf/emnlp/Ethayarajh19,DBLP:conf/nlpir/WangNL20} demonstrate that fastText \cite{bojanowski2017enriching}, Glove \cite{DBLP:conf/emnlp/PenningtonSM14} or their combination outperform BERT on some datasets.

In this paper, we contend that relying on a single pre-trained language model is inadequate for KGE models for two main reasons.
Firstly, the textual information available in different knowledge graphs can vary significantly, which can lead to varying performance of pre-trained language models.
For example, entity descriptions in Freebase may consist of multiple sentences, whereas in Wikidata, they may be just a short sentence. Secondly, different pre-trained language models excel in different levels of information.
For example, while BERT excels in extracting word and sentence-level information, Doc2Vec \cite{DBLP:journals/corr/LeM14} captures the document-level information.
As a result, the use of multiple pre-trained language models is crucial in improving the performance of KGE models.
Therefore, we extend previous KGE models that integrate textual information by incorporating multiple textual representations.
These representations capture different levels of semantics through the use of different pre-trained language models for word, sentence, and document embeddings.
In order to integrate these various representations efficiently, we employ a 4*D dimensional space of hypercomplex numbers to represent structured and textual knowledge of KGs in a unified representation.
We utilize Dihedrons as 4*D dimensional hypercomplex spaces, where each textual representation is a basis and the link prediction process can be modeled by a rotation (from source entity to target entity) in various geometric subspaces induced by Dihedron numbers.
This allows us to model interactions between different textual representations jointly in our KGE model.

Our contributions can be summarized as follows:
\begin{enumerate}
    \item To our knowledge, we are the first work to incorporate multiple textual representations from pre-trained language models into a KGE model.
    \item We develop a novel KGE model with Dihedron algebra, which is more versatile in representing interactions between textual representations from pre-trained language models.
    \item We conduct informative experiments, ablation studies, and analyses on various datasets to investigate the impact of incorporating different pre-trained language models into KGE models.
\end{enumerate}

\section{Related Work}
\subsection{Structural Knowledge Graph Embedding Models}
We will begin by introducing KGE models that consider only structural information from knowledge graphs. TransE \cite{bordes2013translating} and TransH \cite{wang2014b_knowledge} are early KGE models that score triples using different distance-based scoring functions.
These models minimize the distance between the tail entity and the query, which is formulated by adding the head entity and the relation (edge) label.
TransH \cite{wang2014b_knowledge} further improves on TransE by projecting entities onto relation-specific hyperplanes, resulting in different representations of an entity with respect to different relation labels. ComplEx \cite{trouillon2016complex} uses the Hermitian dot product in complex space to model the scoring function, which characterizes the asymmetric relations in knowledge graphs.
RotatE \cite{DBLP:conf/iclr/SunDNT19} is more expressive than previous approaches, as it can represent a relation composed of the natural join between two other relations.
To achieve this, it models relations using rotations in the complex space.
AttE and AttH~\cite{DBLP:conf/acl/ChamiWJSRR20} incorporate an attention mechanism on rotation and reflection in hyperbolic space.
QuatE \cite{Zhang2019QuaternionKG} and Dihedral \cite{DBLP:conf/acl/XuL19,nayyeri2022dihedron} model KGEs using Quaternions and Dihedrons, respectively.  
While these models rely on structural knowledge in knowledge graphs, they do not exploit the advantages of complementary knowledge such as text.
On the contrary, our method jointly utilizes both structural and textual information of KGs.
As a result, we are better equipped to tackle the incompleteness of KGs.

\subsection{Text-enhanced Knowledge Graph Embedding Models}
Several KGE approaches have been proposed that integrate textual information, such as textual descriptions and names of entities, along with the structural information of KGs.
As an early work, DKRL~\cite{xie2016DKRL} extends TransE by considering entity descriptions.
Entity descriptions are encoded by a Convolutional Neural Network (CNN) and jointly optimized together with the TransE scoring function.
ConMask \cite{shi2017open} claims that an entity may involve multiple relations.
It extracts relation-specific information from the entity description by training a network that computes attention over entity descriptions.
However, these approaches were proposed before the emergence of large-scale pre-trained language models like BERT, and as a result, their performance is limited.

Recent approaches have exploited large-scale pre-trained language models to enhance the performance of KGE models.
KG-BERT \cite{PLMyao2019kgbert} and LAnguage Model Analysis (LAMA) \cite{petroni-etal-2019-language} demonstrate that structural knowledge in KGs can be stored in pre-trained language models.
They treat triples in the KG as a sequence of text and obtain the representation of triplets with BERT.
PretrainKGE \cite{zhang-etal-2020-pretrain} extracts knowledge from BERT by representing entity descriptions and relation names with BERT embeddings, which can be utilized to enhance different KGE algorithms.
MLMLM \cite{DBLP:conf/acl/ClouatreTZC21} claims that KG-BERT cannot generalize well on unseen entities.
This work proposes a novel masked language model that yields strong results with unseen entities of arbitrary length in the link prediction task.
StAR \cite{DBLP:conf/www/WangSLZW021} argues that previous work such as KG-BERT may not be effective in learning the structural knowledge in KGs well, and the evidence is that KG-BERT can achieve good results on Top-K recall when K is relatively large (Top-10) but performs poorly when K is small (Top-1).
To address this issue, StAR proposes a hybrid framework that combines pre-trained language models with KGE methods such as RotatE, which aims to obtain the benefits of both approaches.
More recent approaches \cite{PLMli2021siamese,DBLP:conf/coling/Shen0GS22,DBLP:conf/acl-deelio/BrayneWC22} also explore different methods to combine pre-trained language models and KGE to improve the performance of link prediction tasks in KGs.
However, all the KGE approaches mentioned above only incorporate one pre-trained language model, which may not be optimal for all KGs.
Evaluations \cite{DBLP:conf/emnlp/Ethayarajh19,DBLP:conf/nlpir/WangNL20} have shown that pre-trained language models perform differently on different datasets, indicating that there is no single optimal pre-trained language model for all KGs.
In contrast, our approach incorporates multiple pre-trained language models.
This allows us to not only capture different levels of information (word/sentence/document) in a single text but also better fit the different textual information present in various KGs.

\section{Preliminaries}
In this section, we introduce the preliminaries necessary to understand our proposed models.

\paragraph{\textbf{Knowledge Graph:}} A knowledge graph is a collection of triples $\mathcal{K} = \{(h,r,t) |$ \\
$ h,t \in \mathcal{E}, r \in \mathcal{R} \} \subset \mathcal{E} \times \mathcal{R} \times \mathcal{E}$, where $\mathcal{E}$ and $\mathcal{R}$ are the sets of all entities and relation labels in the KG, respectively.

\paragraph{\textbf{Textual Knowledge Graph:}} For a given KG $\mathcal{K}$, we can collect words, sentences, or documents associated with each node or relation label to construct a textual KG $\mathcal{TK}$ defined as $\mathcal{TK} = \{(h^T,r^T,t^T) | h^T,t^T \in \mathcal{E}^T, r^T \in \mathcal{R}^T\} \subset \mathcal{E}^T \times \mathcal{R}^T \times \mathcal{E}^T$, where $h^T, r^T, t^T$ denote the word ($T=W$), sentence ($T=S$), or document ($T=D$) representations of entities and relation labels.
For instance, consider an entity $h = $ \textit{``Berlin''} in the KG that has the word representation $h^W = $ \textit{``Berlin''}, sentence representation $h^S =$ \textit{``Berlin is the capital and largest city of Germany by both area and population''}, and document representation $h^D =$ \textit{``Berlin is the capital and largest city of Germany by both area and population. Its 3.7 million inhabitants make it the most populous city in the European Union...''}.

\paragraph{\textbf{Knowledge Graph Embedding:}} A knowledge graph $\mathcal{K}$ can be represented by a low-dimensional vector embedding, denoted as $\mathcal{KGE} = \{(\mathbf{h,r,t}) | \mathbf{h,t} \in \mathbf{E}, \mathbf{r} \in \mathbf{R}\} \subset \mathbf{E} \times \mathbf{R} \times \mathbf{E}$, where $\mathbf{E,R}$ are the sets of entity and relation label embeddings in the KG, respectively.
These embeddings are $n_e \times D$ and $n_r \times D$ dimensional, respectively, where $n_e$ and $n_r$ are the number of entities and relation labels, and $D$ is the embedding dimension.

\paragraph{\textbf{Pre-trained Language Model KG Embedding:}} The word, sentence, and document representations of triples can be vectorized using pre-trained language models. Thus, we represent the embedding of $\mathcal{TK}$ as a set $\mathcal{TKE} = \{(\mathbf{h}^T, \mathbf{r}, \mathbf{t}^T) | \mathbf{h}^T, \mathbf{t}^T \in \mathbf{E}^T, \mathbf{r} \in \mathbf{R}\} \subset \mathbf{E}^T \times \mathbf{R} \times \mathbf{E}^T$, where $\mathbf{E}^T$ represents the embeddings of all word/sentence/document representations of entities in the KG. These embeddings are generated by feeding the word/sentence/document representation of an entity into the corresponding pre-trained language model.

\paragraph{\textbf{Quaternion and Dihedron Algebra:}} To integrate the structural, word, sentence, and document representations of an entity or a relation label into a unified representation, we utilize a 4*D dimensional algebra, which has been extensively studied in the field of hypercomplex numbers. Specifically, we employ Quaternion $\mathcal{Q}$ \cite{Zhang2019QuaternionKG} and Dihedron $\mathcal{D}$ \cite{DBLP:conf/acl/XuL19,toth2002glimpses,nayyeri2022dihedron} algebra as 4*D dimensional hypercomplex numbers, which are defined as $u = s + x \boldsymbol{i} + y \boldsymbol{j} + z \boldsymbol{k}$, where $\boldsymbol{i}, \boldsymbol{j}$, and $\boldsymbol{k}$ are the three imaginary parts.
In Quaternion and Dihedron representations, we have, respectively,
$\boldsymbol{i^2 =\, j^2 = \, k^2 = ijk = \Bar{1}}, \,
    \boldsymbol{ij=k, \, jk = i, \, ki = j},$
   $\boldsymbol{ji = \Bar{k}, \, kj = \Bar{i}, \, ik = \Bar{j}},$ and
    $\boldsymbol{i^2 = \Bar{1}, \, j^2 = k^2 = 1}, \,
    \boldsymbol{ij=k, \, jk = \Bar{i}, \, ki = j},$
    $\boldsymbol{ji = \Bar{k}, \, kj = i, \, ik = \Bar{j}},$
where $\Bar{a} = -a, \, a \in {i,j,k}$.
The prime operators of Quaternion and Dihedron are defined in the following paragraphs.

\textbf{Quaternion Product:} This product is also known as the Hamilton product. 
Let $u = s_u + x_u \boldsymbol{i} + y_u \boldsymbol{j} + z_u \boldsymbol{k}, v = s_v + x_v \boldsymbol{i} + y_v \boldsymbol{j} + z_v \boldsymbol{k}$ be two Quaternion numbers. 
The Hamilton product between $u,v$ is defined as follows:
\begin{equation}
\begin{aligned}
u \otimes_\mathcal{Q} v
    &:=\left(s_u s_v-x_u x_v-y_u y_v-z_u  z_v\right)
+\left(s_u x_v + x_u s_v  +y_u z_v-z_u  y_v\right) \mathbf{i} \\
&+\left(s_u  y_v-x_u  z_v+y_u  s_v+z_u  x_v\right) \mathbf{j}
+\left(s_u  z_v+x_u  y_v-y_u  x_v+z_u  s_v\right) \mathbf{k}.
\label{quatproduct}
\end{aligned}
\end{equation}

\textbf{Dihedron Product:} Let $u,v$ be two Dihedron numbers. The Dihedron product is defined as
\begin{equation}
\begin{aligned}
u \otimes_\mathcal{D} v
    &:=\left(s_u  s_v-x_u x_v+y_u  y_v+z_u  z_v\right)
+\left(s_u  x_v+x_u  s_v-y_u  z_v+z_u  y_v\right) \mathbf{i} \\
&+\left(s_u  y_v-x_u  z_v+y_u  s_v+z_u  x_v\right) \mathbf{j} 
+\left(s_u  z_v+x_u  y_v-y_u  x_v+z_u  s_v\right) \mathbf{k}.
\label{dihedronproduct}
\end{aligned}
\end{equation}

\textbf{Inner Product:} In both the Dihedron and the Quaternion spaces, the inner product is defined as $\langle u, v \rangle = u \cdot v :=s_u  s_v + x_u  x_v + y_u  y_v + z_u  z_v.$

\textbf{Conjugate:} The conjugate in both representations is $\Bar{u} = s - x \boldsymbol{i} - y \boldsymbol{j} - z \boldsymbol{k}.$

\textbf{Norm:} The norm in the representations of the Quaternions and Dihedrons is defined as $\| u \| = \langle u, \Bar{u} \rangle$ which are $\sqrt{s^2_u+ x^2_u + y^2_u + z^2_u}$ and $\sqrt{s^2_u+ x^2_u - y^2_u - z^2_u}$, respectively.

Previous approaches \cite{Zhang2019QuaternionKG,DBLP:conf/aaai/CaoX0CH21} employed Quaternion as another algebra in the hypercomplex space.
Both the Quaternion and Dihedron spaces offer various geometric representations.
Quaternion numbers of equal length represent hyperspheres, while Dihedron numbers of equal length can represent various shapes, including spheres, one-sheets, two-sheets, and conical surfaces.
Therefore, the Dihedron space is more expressive than the Quaternion space in terms of geometric representation.

\section{Our Method}
\label{sec:method}
\begin{figure*}[ht]
    \centering
    \includegraphics[width=10cm,height=5cm]{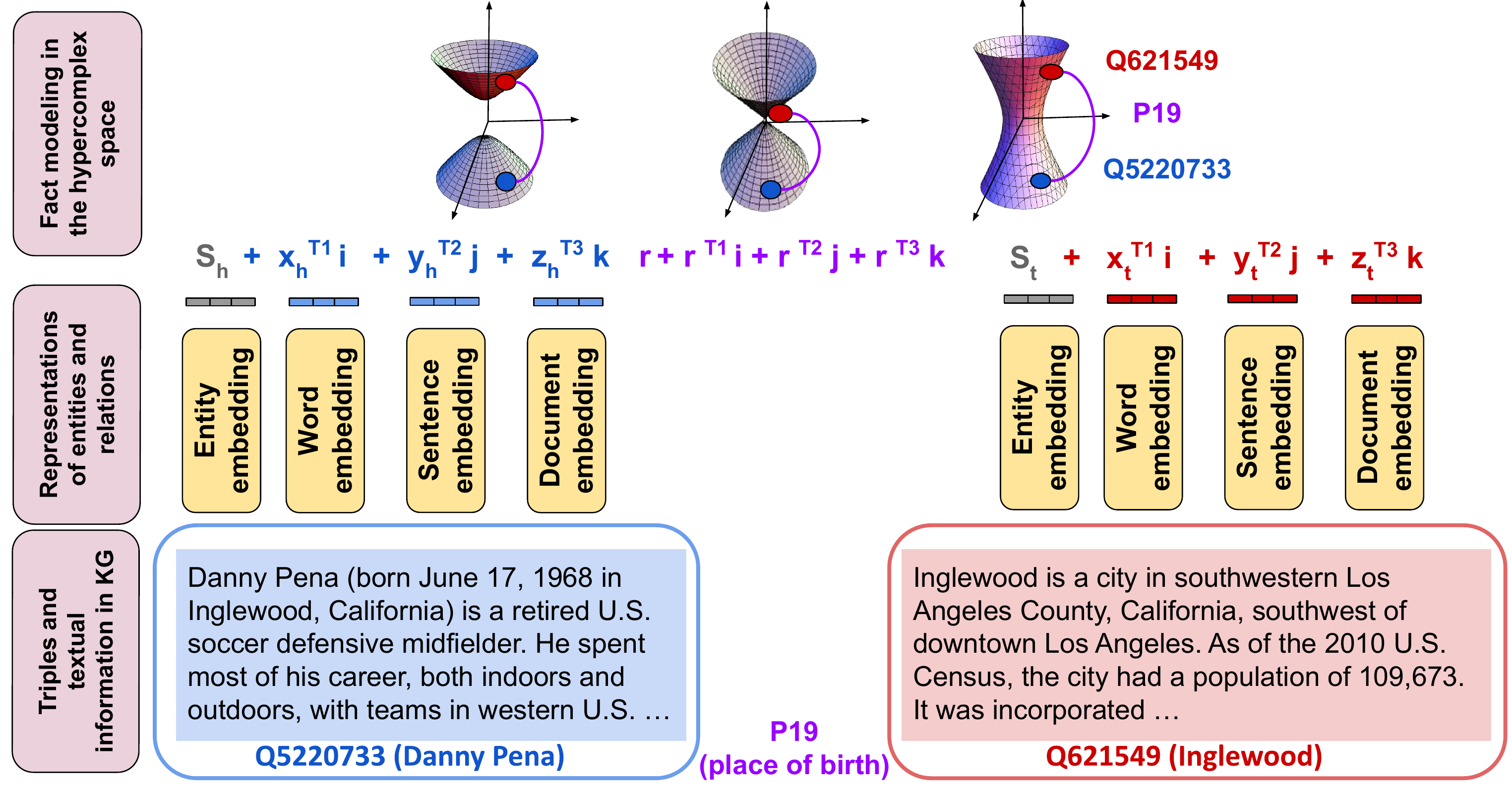}
    \caption{The proposed model's overall architecture considers various entity representations, such as word, sentence, and document levels, and maps them into a joint geometric space using the Quaternion or Dihedron algebra. This joint space can be either spherical or hyperbolic.}
    \label{fig: architecture}
\end{figure*}

In this section, we introduce a family of embedding models based on the Quaternion or Dihedron algebra that operate in 4*D dimensional spaces.
These spaces capture the four types of entity representations, namely graph, word, sentence, and document embeddings, and allow for modeling the interactions between different elements, resulting in a comprehensive feature representation that can be used to assess the plausibility score of links.
As shown in Figure~\ref{fig: architecture}, our framework involves splitting a triple, such as (Q5220733, P19, Q621549), into a triple pattern, e.g., (Q5220733, P19, ?), and a tail, e.g., Q621549, which corresponds to a query such as ``Where was Danny Pena born?''.
We then compute embeddings for the query and the tail using both textual and structural information, and calculate the plausibility score by measuring the distance between them in the embedding space.

In the rest of this section, we will present our embedding model and provide technical details on the dimension adjustment that aligns the different representations of entities in the same geometric space.
This process is accomplished through the following order:
a) Entity, Relation, and Query Representation,
b) Dimension Adjustment and
c) Triple Plausibility Score and Training.
The purpose of b) is to match the dimensions of the KGE model with the pre-trained vectors from language models in cases where the dimensions do not match.

\subsection{Entity, Relation, and Query Representation}
We represent each entity $e$ as a Quaternion or Dihedron vector with dimensions $4*D$, which captures its structural and textual information at different levels of granularity, as follows:
\begin{equation}
    \mathbf{e} = \mathbf{s}_e + \mathbf{x}^{\mathcal{T}_1}_e \boldsymbol{i} + \mathbf{y}^{\mathcal{T}_2}_e \boldsymbol{j} + \mathbf{z}^{\mathcal{T}_3}_e \boldsymbol{k},
    \label{entityrep1}
\end{equation}
where $\mathbf{s}_e$ is the node representation in the graph embedding of the KG and $\mathcal{T}_i, i=1,2,3$ are pre-trained language models.
For one entity description, we feed the text into a set of pre-trained language models $\mathcal{T}_i$ and initialize corresponding embeddings $\mathbf{x}^{\mathcal{T}_1}_e, \mathbf{y}^{\mathcal{T}_2}_e, \mathbf{z}^{\mathcal{T}_3}_e$.

In Figure~\ref{fig: baseMotivExamp}, the entities ``Q5220733'' and ``Q621549'' both possess textual descriptions. 
To extract word, sentence, and document representations from these descriptions, we use a variety of pre-trained language models.
By doing so, we obtain a comprehensive representation of the entities.

Each relation is represented as a rotation in the hypercomplex space, which is  
\begin{equation}
    \mathbf{r} = \frac{\mathbf{s}_r + \mathbf{x}^{\mathcal{T}_1}_r \boldsymbol{i} + \mathbf{y}^{\mathcal{T}_2}_r \boldsymbol{j} + \mathbf{z}^{\mathcal{T}_3}_r \boldsymbol{k}}{ \sqrt{\mathbf{s}^2_r + (\mathbf{x}^{\mathcal{T}_1}_r)^2 + (\mathbf{y}^{\mathcal{T}_2}_r)^2 + (\mathbf{z}^{\mathcal{T}_3}_r)^2}} .
\end{equation}

To calculate the plausibility score of a triple $(h,r,t)$, we divide it into a triple pattern $(h,r,?)$ and a tail entity $t$.
Both the triple pattern and the tail entity are represented in a 4*D dimensional space.
Table~\ref{tbl:queries} presents three different approaches for representing the triple pattern $(h,r,?)$ in our model.
The following methods are proposed for the representation of the query:
\begin{table*}[h!]
\centering
\begin{adjustbox}{width=\textwidth}
\begin{tabular}{c|c|c}
\hline
Name & query $(h,r,?)$ & embedding\,\,$\mathbf{h,r} \in \mathcal{D}^d$ \\ 
\hline
Tetra & $\mathbf{q} = \mathbf{h} \otimes \mathbf{r}^{\triangleleft}$ & $\mathbf{h} = \mathbf{s}_h + \mathbf{x}^{\mathcal{T}_1}_h \boldsymbol{i} + \mathbf{y}^{\mathcal{T}_2}_h \boldsymbol{j} + \mathbf{z}^{\mathcal{T}_3}_h \boldsymbol{k}, \mathbf{r}^{\triangleleft} = \frac{\mathbf{r}}{|\mathbf{r}|}$ \\
\hline
Robin & $\mathbf{q} = \mathbf{h} \otimes \boldsymbol{h}_{lt}^{\triangleleft^{\mathcal{T}_1}} + \boldsymbol{h}_l^{\mathcal{T}_0} + \boldsymbol{h}^{\mathcal{T}_1}_t + \mathbf{r}^{\triangleleft} $ & $ \boldsymbol{h}_{lt}^{\mathcal{T}_1} = \mathbf{s}^{l^{\mathcal{T}_0}}_h + \mathbf{x}^{l^{\mathcal{T}_0}}_h \boldsymbol{i} + \mathbf{y}^{t^{\mathcal{T}_1}}_h \boldsymbol{j} + \mathbf{z}^{t^{\mathcal{T}_1}}_h \boldsymbol{k}, \boldsymbol{h}_{lt}^{\triangleleft^{\mathcal{T}_1}} = \frac{\boldsymbol{h}_{lt}^{\mathcal{T}_1}}{|\boldsymbol{h}_{lt}^{\mathcal{T}_1}|}$\\
\hline
 Lion & $\mathbf{q} = \mathbf{h} \otimes \boldsymbol{h}_{t}^{\triangleleft^{\mathcal{T}_1\mathcal{T}_2}} + \boldsymbol{h}^{\mathcal{T}_1}_{t} + \boldsymbol{h}^{\mathcal{T}_2}_{t} + \mathbf{r}^{\triangleleft} $ & $ \boldsymbol{h}_{t}^{\mathcal{T}_1\mathcal{T}_2} = \mathbf{s}^{t^{\mathcal{T}_1}}_h + \mathbf{x}^{t^{\mathcal{T}_1}}_h \boldsymbol{i} + \mathbf{y}^{t^{\mathcal{T}_2}}_h \boldsymbol{j} + \mathbf{z}^{t^{\mathcal{T}_2}}_h \boldsymbol{k}, \boldsymbol{h}_{t}^{\triangleleft^{\mathcal{T}_1\mathcal{T}_2}} = \frac{\boldsymbol{h}_{t}^{\mathcal{T}_1\mathcal{T}_2}}{|\boldsymbol{h}_{t}^{\mathcal{T}_1\mathcal{T}_2}|}$\\
\hline
\end{tabular}
\end{adjustbox}
\caption{Query representations derived by our models.}
\label{tbl:queries}
\end{table*}

\paragraph{\textbf{Text-enhanced relaTional RotAtion (Tetra)}:}
This model comprises four parts for representing each entity.
For each head entity, node representation $\mathbf{s}_h$ is learned from a graph embedding of the KG structure, and the components ($\mathbf{\mathbf{x}^{\mathcal{T}_1}_h}, \mathbf{y}^{\mathcal{T}_2}_h, \mathbf{z}^{\mathcal{T}_3}_h$) are learned from textual embeddings (word, sentence, and document) of the entities using three pre-trained language models.
Likewise, for each tail entity, $\mathbf{s}_t$ and ($\mathbf{\mathbf{x}^{\mathcal{T}_1}_t}, \mathbf{y}^{\mathcal{T}_2}_t, \mathbf{z}^{\mathcal{T}_3}_t$) are also learned.
The Tetra model represents the query by learning relation-specific rotations of the head in Quaternion or Dihedron space, and $\otimes$ refers to the Quaternion or Dihedron products (see equation \ref{quatproduct}, \ref{dihedronproduct}) in Quaternion or Dihedron space.
The resulting query inherently contains pairwise correlations between each of $\mathbf{s}_h, \mathbf{\mathbf{x}^{\mathcal{T}_1}_h}, \mathbf{y}^{\mathcal{T}_2}_h, \mathbf{z}^{\mathcal{T}_3}_h$, thus providing a rich feature representation for the corresponding query.

\paragraph{\textbf{Multi-texts RelatiOn-Based rotatIon and translatioN (Robin)}:}
In this model, we incorporate both the vector of the entity name (e.g., ``Berlin'') indexed by $l$ (i.e., $\boldsymbol{h}^{\mathcal{T}_0}_l$) and the vector of the textual description indexed by $t$ (i.e., $\boldsymbol{h}^{\mathcal{T}_1}_t$).
We perform a rotation of the form $\boldsymbol{h} \otimes \boldsymbol{h}_{lt}^{\triangleleft^{\mathcal{T}1}}$ and a translation of the form $\boldsymbol{h}^{\mathcal{T}_0}_l + \boldsymbol{h}^{\mathcal{T}_1}_t + \boldsymbol{r}^{\triangleleft}$ derived from these two sources of information.
Specifically, $\boldsymbol{h}_{lt}^{\triangleleft^{\mathcal{T}_1}}$ is obtained by normalizing $\boldsymbol{h}^{\mathcal{T}1}_{lt}$.
We always use Word2Vec ($\mathcal{T}_0$) to embed entity names, but we use a different pre-trained language model for embedding entity descriptions, which is selected based on performance in experiments (details provided in the experimental section).
We utilize two distinct neural networks (details in Section \ref{Sec:DimensionAdjustment}) to generate distinct representations $\mathbf{s}^{l^{\mathcal{T}_0}}_h$ and $\mathbf{x}^{l^{\mathcal{T}_0}}_h$ given the same embedding from Word2Vec $\mathcal{T}_0$.
Similarly, $\mathbf{y}^{t^{\mathcal{T}_1}}_h$ and $\mathbf{z}^{t^{\mathcal{T}_1}}_h$ are distinct representations given the same embedding from another pre-trained language model $\mathcal{T}_1$.
Consequently, the query is computed by combining the head entity embedding from the KG, relation label embedding, entity name, and entity description through translation and rotation in Quaternion or Dihedron space:
$\mathbf{q} = \boldsymbol{h} \otimes \boldsymbol{h}_{lt}^{\triangleleft^{\mathcal{T}_1}} + \boldsymbol{h}^{\mathcal{T}_0}_l + \boldsymbol{h}^{\mathcal{T}_1}_t + \mathbf{r}^{\triangleleft},$
where $h$ is the graph embedding of the head entity.
Incorporating both translation and rotation improves the model's expressiveness compared to utilize only a single rotation~\cite{nayyeri2020fantastic,DBLP:conf/acl/ChamiWJSRR20}.

\paragraph{\textbf{Multi-Language models relatIon rOtation and translatioN (Lion)}:}
In this model, we utilize two pre-trained language models ($\mathcal{T}_1, \mathcal{T}_2$) to embed the same entity description and adjust their dimensions using NNs (detailed in Section \ref{Sec:DimensionAdjustment}). We then construct a vector for rotation as follows:
$ \boldsymbol{h}^{\mathcal{T}_1\mathcal{T}_2} = \mathbf{s}^{t^{\mathcal{T}_1}}_h + \mathbf{x}^{t^{\mathcal{T}_1}}_h \boldsymbol{i} + \mathbf{y}^{t^{\mathcal{T}_2}}_h \boldsymbol{j} + \mathbf{z}^{t^{\mathcal{T}_2}}_h \boldsymbol{k}, {\boldsymbol{h}^{\mathcal{T}_1\mathcal{T}_2}}^{\triangleleft} = \frac{\boldsymbol{h}^{\mathcal{T}_1\mathcal{T}_2}}{|\boldsymbol{h}^{\mathcal{T}_1\mathcal{T}_2}|}$.
Together with the rotation $\mathbf{h} \otimes \boldsymbol{h}_{t}^{\triangleleft^{\mathcal{T}_1\mathcal{T}_2}}$, we use the embedding of the entity description $\boldsymbol{h}^{\mathcal{T}_1}_{t}, \boldsymbol{h}^{\mathcal{T}_2}_{t}$ to represent a translation for query representation in Quaternion or Dihedron spaces: 
$\mathbf{q} = \boldsymbol{h} \otimes \boldsymbol{h}_t^{\triangleleft^{\mathcal{T}_1\mathcal{T}_2}} + \boldsymbol{h}_t^{\mathcal{T}_1} + \boldsymbol{h}_t^{\mathcal{T}_2} + \mathbf{r}^{\triangleleft}.$

The query representations mentioned above are related to the triple pattern $(h,r,?)$.
For the triple pattern $(?, r, t)$, we adopt the approach of using $(t, r^{-1}, ?)$ for query representation, where $r^{-1}$ is a reverse relation label corresponding to the relation label $r$.
To create a set of $|R|$ embeddings of reverse relation labels $R^{-1}$, we add an additional triple $(t, r^{-1}, h)$ to the training set for each embedding of the reverse relation label $r^{-1} \in R^{-1}$, following the approach used in previous works \cite{kazemi2018simple,lacroix2018canonical}.
When representing queries for these triples, we use the equations presented in Table~\ref{tbl:queries}, but with $h$ replaced by $t$ and $r$ replaced by $r^{-1}$.
It is worth noting that the choice of the appropriate model from Table~\ref{tbl:queries} may depend on various characteristics of the KG, such as sparsity and density, quality of textual description, etc.
For instance, if the KG has a complex structure, a more expressive model like the Robin and Lion may be required,
because they mix both the translation and the rotation, which could be preferred over the Tetra relying solely on the rotation.

\subsection{Dimension Adjustment}
\label{Sec:DimensionAdjustment}
The underlying assumption of Equation~\ref{entityrep1} is that the vectors $\mathbf{s}_e, \mathbf{x}_e^{\mathcal{T}_1}, \mathbf{y}_e^{\mathcal{T}_2},$ and $\mathbf{z}_e^{\mathcal{T}_3}$ have the same dimension $D$.
However, since pre-trained language models may produce vectors of different dimensions, we use a neural network to adjust their dimensions. Thus, we can rewrite Equation~\ref{entityrep1} as follows:
\begin{equation}
    \mathbf{e} = \mathbf{s}_e + NN(\mathbf{x}^{\mathcal{T}_1}_e) \boldsymbol{i} + NN(\mathbf{y}^{\mathcal{T}_2}_e) \boldsymbol{j} + NN(\mathbf{z}^{\mathcal{T}_3}_e) \boldsymbol{k},
    \label{entityrep}
\end{equation}
where $NN$ is a multilayer perceptron whose input and output dimensions are $D_{\mathcal{T}_i}$ and $D$, respectively.

In our dimension-adjustment module, we use an individual multi-layer network $NN$ for each embedding from pre-trained language models.
The input and output dimension of each $NN$ are $D_{\mathcal{T}_i}$ and  $D$, respectively.
Among all embeddings, the largest embedding from BERT has size of 512, so we always adapt all embedding dimension $D$ to 512.
Our $NN$ has two layers, each layer consists of a linear transformation and then a non-linear transformation.
For the activation function in the non-linear transformation, we simply use hyperbolic tangent function.

\subsection{Triple Plausibility Score and Training}
To measure the plausibility score of a triple $(h,r,t)$, we calculate the distance between the query $\mathbf{q}$ and the corresponding tail $\boldsymbol{t}$ as follows:
\begin{equation}
    f(h,r,t) = -d(\boldsymbol{q}, \boldsymbol{t}) + b_h + b_t,
\end{equation}
where $b_h, b_t$ are entity-specific biases proposed in~\cite{balazevic2019multi}.

In Figure~\ref{fig: baseMotivExamp}, a series of geometric representations are presented (located on top of the figure) in which the query and corresponding tail are matched (from the blue dot to the red dot).
The textual description of ``Q5220733'' mentions the entity ``Q621549'' and the term ``born'', which is closely related to the ``P19'' relation label.
Similarly, the textual description of ``Q621549'' includes the entities ``Inglewood''.
These descriptions are strongly correlated and effectively cover the mention of the triple elements (head, relation label, tail).
Consequently, based on the textual descriptions of these entities, it can be inferred that ``Q5220733'' was born in ``Inglewood''.

During the training phase, we minimize the cross-entropy loss function with uniform negative sampling as described in \cite{DBLP:conf/acl/ChamiWJSRR20}.
In particular, for a positive triple $(h, r, t)$, we obtain negative triples by uniformly sampling negative tail entities $t' \in E$ such that $(h, r, t') \notin \mathcal{K}$. 
Besides, for any positive triple, we always train another triple involving inverse relation $(t, r^{-1}, h')$, where we similarly perform negative sampling to obtain negative head entities $h'$.
This is based on previous work \cite{DBLP:conf/icml/LacroixUO18} that suggests the inclusion of triples with inverse relations can improve model performance.

Given positive and negative triples, the loss function can be defined as follows:
\begin{equation}
    \mathcal{L} = - \sum_{h, r, t ; t'} y^{t ; t^{'}} log f(h, r, t ; t') - \sum_{t, r^{-1}, h ; h'} y^{h ; h^{'}} log f(t, r^{-1}, h ; h')
    \label{Eq.Loss}
\end{equation}
where $y^{t ; t^{'}} = \pm 1$ is the label for positive or negative triples when sampling tail entities, and $y^{h ; h^{'}}$ is similarly for head entities.
During the training process, we employ the Adagrad optimizer \cite{DBLP:journals/jmlr/DuchiHS11} to optimize the model parameters. Furthermore, we utilize early stopping on the validation dataset to prevent overfitting.


\section{Experiments}
\subsection{Experimental Setup}
\paragraph{\textbf{Datasets, Environments and Hyperparameters:}}
We evaluate our proposed models on two domain-specific KG datasets: \textbf{NATIONS} \cite{kok2007statistical}, \textbf{Diabetes} \cite{10.1093/bioinformatics/btac085}, and two commonsense KG datasets: \textbf{FB15k-237} \cite{DBLP:conf/acl-cvsc/ToutanovaC15}, and \textbf{YAGO-10}.
Table~\ref{tab:Datasets statistics} provides a summary of types and other statistic details of each dataset.
Other details of datasets can be found in Appendix A.
Besides, the details of our environments and hyperparameters can be found in Appendix B.

\begin{table}[!h]
\centering
\begin{tabular}{|c|c|c|c|c|c|c|}
\hline
\textbf{Dataset} & Type & \textbf{\#ent} & \textbf{\#rel} & \textbf{\#train} & \textbf{\#val} & \textbf{\#test} \\
\hline
NATIONS & domain specific & 14 & 55 & 1,592 & 199 & 201\\
Diabetes & domain specific & 7,886 & 67 & 56,830  & 1,344 &  1,936\\
FB15k-237 & commonsense & 14,904 & 237 & 271,431 & 17,503 & 20,427 \\
YAGO-10  & commonsense & 103,222 & 30 & 490,214& 2,295 & 2,292\\
\hline
\end{tabular}
\caption{The statistics of our datasets.}
\label{tab:Datasets statistics}
\end{table}

\paragraph{\textbf{Evaluation Metrics:}}
We evaluated our models using the link prediction task and the following standard evaluation metrics: Mean Reciprocal Rank (MRR) and Hits@K, where K is set to 1, 3, and 10.
MRR calculates the mean reciprocal rank of the correct tail entity across all queries, while Hits@K measures the proportion of correct tail entities that rank in the top K positions.
In accordance with prior work, we also employed the filtering setup \cite{NIPS2013_1cecc7a7} during evaluation to remove existing triples in the dataset from the ranking process.

\paragraph{\textbf{Baselines and Our Ablation Models:}}
We compare our proposed models against four baselines that do not consider textual information: TransE \cite{bordes2013translating}, ComplEx \cite{trouillon2016complex}, AttE, and AttH \cite{DBLP:conf/acl/ChamiWJSRR20}.
We also compare against baselines that incorporate textual information: DKRL~\cite{xie2016DKRL}, ConMask \cite{shi2017open}, PretrainKGE \cite{zhang-etal-2020-pretrain}, KG-BERT \cite{PLMyao2019kgbert}, and StAR \cite{DBLP:conf/www/WangSLZW021}.
We reimplemented ConMask and PretrainKGE and obtained the codes for DKRL, KG-BERT and StAR online.

We conducted an ablation study to evaluate different variants of our model with different combinations of pre-trained language models.
To simplify, we use the abbreviations $\{W, F, D, S\}$ to represent the pre-trained language models {$W$ord2Vec, $F$astText, $D$oc2Vec, $S$entenceTransformer}, respectively.
We compared four variants of Robin that utilize one of $\{W, F, S, D\}$ individually for modeling the entity descriptions and always use $W$ for modeling the entity names, namely, \textit{Robin\_W}, \textit{Robin\_F}, \textit{Robin\_S} and \textit{Robin\_D}.
We further compared two variants of Lion that use $\{S, D\}$ or $\{F, S\}$, namely, \textit{Lion\_SD} and \textit{Lion\_SF}.
We also compare three variants of Tetra that use only $\{S, F\}$ for \textit{Tetra\_SF} and use $\{W, S, F\}$ for \textit{Tetra\_WSF}.
Besides, we do not incorporate any pre-trained language model into \textit{Tetra\_zero}.
All our variants employ the Dihedron representation for better performance.
By comparing the performance of our ablation models, we evaluated different aspects of our model, as shown in Table 3:

\begin{table}
\centering
\begin{adjustbox}{width=\textwidth,center}
    \begin{tabular}{|c|c|}
    \hline
      Ablation models to compare  & Purpose of evaluation \\
      \hline
    different variants of Robin  & difference from each individual pre-trained language model \\
    \hline
    different variants of Lion  & different combinations of two pre-trained language models \\
    \hline
    different variants of Tetra & difference from numbers of incorporated pre-trained language models \\
    \hline
    Robins and Lions & different numbers of texts (name, description) utilized in our model \\
    \hline
    \end{tabular}
\end{adjustbox}
\label{Tab.Ablation}
\caption{The design of our ablation models.}
\end{table}

\subsection{Link Prediction results and analysis}
Table~\ref{tab:results for Nations, Diabetes, FB15k-237, YAGO-10} presents the results of the link prediction task for the embedding dimensions $D=32$ and $D=500$, respectively.
Our proposed models consistently outperform all baselines on all datasets.

However, the performance of our ablation variants varies across different datasets.
In the small NATIONS dataset, Tetra outperforms Robin and Lion by a significant margin.
This is due to the fact that on smaller datasets, the amount of structural information from the KG is limited, and the incorporation of additional textual information becomes more crucial.
Tetra performs better in this scenario because it incorporates three different pre-trained language models, allowing for a more comprehensive exploitation of textual information.
The results on the three KGs suggest that Lion generally outperforms Robin, with the most significant improvement of a 3\% higher MRR observed on YAGO-10 in the low-dimensional setting.
These results demonstrate that incorporating more pre-trained language models to extract information from the same text is more effective than incorporating more types of text, such as entity names and descriptions. Additionally, Tetra's performance is comparable to Lion in the high-dimensional setting on FB15k-237, with Tetra\_SF even achieving the best performance among all ablation variants on YAGO-10 in the high-dimensional setting.
Since the primary difference between the Tetra and Lion variants is the mixture of translation and rotation, these results suggest that this mixture is more suitable in the low-dimensional setting.

We also compared all the different variants of Robin and Lion models together since they all use two pre-trained language models.
We observed that their performances on smaller datasets such as NATIONS and Diabetes were quite similar, but on larger datasets, there were more noticeable differences.
For instance, on the YAGO-10 dataset, Lion\_SD outperformed Robin\_S by 3.4\% in terms of MRR in the low dimensional setting, and Lion\_SF had 9\% higher MRR than Robin\_D in the high dimensional setting.
These results suggest that the selection of the incorporated pre-trained language models is more critical on larger datasets.

For various Tetra variants, our findings indicate that Tetra\_zero consistently underperforms, which highlights the significance of incorporating pre-trained language models.
Comparing Tetra\_SF and Tetra\_WSF, we observe that the former generally exhibits superior performance on smaller datasets, while the latter performs better on larger datasets.
This evidence suggests that incorporating more pre-trained language models is more appropriate when the structural information is insufficient, such as on smaller datasets.

\begin{table*}[h!]
\centering
\begin{adjustbox}{width=\textwidth,center}
\begin{tabular}{|c|l|cccc|cccc|cccc|cccc|}
\hline
\multirow{2}{*}{\textbf{Elements}} & \multirow{2}{*}{\textbf{Model}} & \multicolumn{4}{c|}{\textbf{Nations}} & \multicolumn{4}{c|}{\textbf{Diabetes}} & \multicolumn{4}{c|}{\textbf{FB15k-237}} & \multicolumn{4}{c|}{\textbf{YAGO-10}} \\ \cline{3-18}
&  & MRR & H@1 & H@3 & H@10 & MRR & H@1 & H@3 & H@10 & MRR & H@1 & H@3 & H@10 &  MRR & H@1 & H@3 & H@10 \\ \hline
\multirow{7}{*}{\makecell[c]{Baselines\\D=32}} &TransE & 0.684 & 0.542 & 0.779 & 0.990 & 0.166  & 0.089 & 0.182 & 0.322 & 0.274 & 0.197 & 0.298 & 0.428 & 0.368 & 0.284 & 0.403 & 0.534 \\
& ComplEx & 0.610 & 0.460 & 0.697 & 0.978 & 0.136 & 0.069 & 0.144 & 0.273 & 0.250 & 0.178 & 0.275 & 0.395 & 0.344 & 0.277 & 0.365 & 0.480 \\
&AttE & 0.648 & 0.488 & 0.741 & 0.980 & 0.125 & 0.060 & 0.135 & 0.259 & 0.283 & 0.205 & 0.307 & 0.436 & 0.364 & 0.289 & 0.394 & 0.518 \\
&AttH & 0.728 & 0.610 & 0.804 & 0.990 & 0.120  & 0.058 & 0.124  & 0.247 & 0.280 & 0.200 & 0.307 & 0.443 & 0.380 & 0.300 & 0.415 & 0.538 \\
& DKRL & 0.660 & 0.505 & 0.774 & \cellcolor{cyan!25}\textbf{0.998} & 0.158 & 0.085 & 0.171 & 0.310 & 0.230 & 0.159 & 0.250 & 0.368 & 0.339 & 0.255 & 0.373 & 0.509 \\
& ConMask & 0.662 & 0.505 & 0.761 & 0.988 & 0.155 & 0.083 & 0.169 & 0.305 & 0.245 & 0.171 & 0.268 & 0.390 & 0.362 & 0.294 & 0.389 & 0.504 \\
& PretrainKGE & 0.674 & 0.540 & 0.756 & 0.985 & 0.151 & 0.078 & 0.164 & 0.300 & 0.251 & 0.175 & 0.276 & 0.397 & 0.349 & 0.274 & 0.380 & 0.502 \\
\hline
\multirow{9}{*}{\makecell[c]{Ours\\D=32}} &Robin\_W & 0.730 & 0.610 & 0.801 & 0.990 & 0.173 & 0.096 & 0.188 & 0.333 & 0.290 & 0.208 & 0.317 & 0.449 & 0.363 & 0.281 & 0.396 & 0.528\\
&Robin\_F & 0.732 & 0.609 & 0.811 & 0.993 & 0.173 & 0.095 & 0.186 & 0.338 & 0.300 & 0.213 & 0.329 & 0.471 & 0.365 & 0.285 & 0.397 & 0.524\\
&Robin\_S & 0.732 & 0.612 & 0.799 & 0.980 & 0.173 & 0.095 & 0.188 & 0.333 & \cellcolor{cyan!25}\textbf{0.304} & \cellcolor{cyan!25}\textbf{0.222} & 0.331 & 0.465 & 0.363 & 0.282 & 0.398 & 0.528\\
&Robin\_D & 0.728 & 0.610 & 0.789 & 0.993 & 0.173 & 0.097 & 0.187 & 0.333 &  0.294 & 0.213 & 0.321 & 0.452 & 0.366 & 0.286 & 0.401 & 0.528\\
& Lion\_SD & 0.736 & 0.624 & 0.801 & 0.988 & 0.167 & 0.090  & 0.180  & 0.330 & \cellcolor{cyan!25}\textbf{0.304} & 0.217 & \cellcolor{cyan!25}\textbf{0.333} & \cellcolor{cyan!25}\textbf{0.478} & \cellcolor{cyan!25}\textbf{0.397} & \cellcolor{cyan!25}\textbf{0.314} & \cellcolor{cyan!25}\textbf{0.441} & \cellcolor{cyan!25}\textbf{0.554} \\
&Lion\_SF & 0.727 & 0.605 & 0.801 & 0.993 & \cellcolor{cyan!25}\textbf{0.175} & \cellcolor{cyan!25}\textbf{0.097} & \cellcolor{cyan!25}\textbf{0.190} & \cellcolor{cyan!25}\textbf{0.340} & 0.301 & 0.214 & 0.330 & 0.475 & 0.395 & \cellcolor{cyan!25}\textbf{0.314} & 0.439 & 0.548 \\
& Tetra\_zero & 0.547 & 0.356 & 0.647 & 0.978 & 0.134 & 0.069 & 0.141 & 0.266 & 0.264 & 0.187 & 0.288 & 0.414 & 0.330 & 0.274 & 0.349 & 0.451 \\
&Tetra\_SF & 0.773 & 0.652 & 0.856 & 0.990 & 0.157 & 0.086 & 0.166 & 0.299 &  0.278 & 0.196 & 0.304 & 0.439 & 0.255 & 0.178 & 0.274 & 0.421 \\
&Tetra\_WSF & \cellcolor{cyan!25}\textbf{0.780} & \cellcolor{cyan!25}\textbf{0.669} & \cellcolor{cyan!25}\textbf{0.858} & 0.995 & 0.155 & 0.084 & 0.169 & 0.302 & 0.266 & 0.188 & 0.289 & 0.421 & 0.169 & 0.113 & 0.180 & 0.288 \\
\hline
\hline
\multirow{7}{*}{\makecell[c]{Baselines\\D=500}} &TransE & 0.712 & 0.590 & 0.789 & 0.990 & 0.178 & 0.100 & 0.194 & 0.341 & 0.318 & 0.231 & 0.350 & 0.492 & 0.421 & 0.351 & 0.461 & 0.556   \\
& ComplEx & 0.626 & 0.483 & 0.699 & 0.978 & 0.144 & 0.077 & 0.155 & 0.283 & 0.308 & 0.223 & 0.337 & 0.482 & 0.410 & 0.341 & 0.443 & 0.550 \\
&AttE & 0.795 & 0.699 & 0.858 & 0.993 & 0.187 & 0.105 & 0.205 & 0.355 & 0.272 & 0.195 & 0.295 & 0.429 & 0.356 & 0.294 & 0.389 & 0.471 \\
&AttH & 0.789 & 0.684 & 0.861 & \cellcolor{cyan!25}\textbf{0.995} & 0.185 & 0.102 & 0.202 & 0.354 & 0.265 & 0.188 & 0.287 & 0.420 & 0.313 & 0.256 & 0.336 & 0.431 \\
& DKRL & 0.706 & 0.582 & 0.786 & 0.990 & 0.162 & 0.083 & 0.176 & 0.328 & 0.239 & 0.169 & 0.260 & 0.375 & 0.333 & 0.239 & 0.371 & 0.520 \\
& ConMask & 0.713 & 0.587 & 0.808 & 0.993 & 0.165 & 0.086 & 0.180 & 0.335 & 0.258 & 0.183 & 0.284 & 0.405 & 0.381 & 0.306 & 0.421 & 0.519 \\
& PretrainKGE & 0.718 & 0.592 & 0.803 & 0.993 & 0.159 & 0.082 & 0.172 & 0.323 & 0.262 & 0.187 & 0.287 & 0.407 & 0.320 & 0.231 & 0.353 & 0.495 \\
\hline
\multirow{9}{*}{\makecell[c]{Ours\\D=500}} &Robin\_W &0.731& 0.614& 0.796& 0.993 & 0.192 & \cellcolor{cyan!25}\textbf{0.111} & 0.207 & 0.363 & 0.328 & 0.235 & 0.362 & 0.514 & 0.402 & 0.327 & 0.442 & 0.541\\
&Robin\_F &0.721& 0.597& 0.791& \cellcolor{cyan!25}\textbf{0.995} & 0.192 & 0.110 & 0.208 & 0.366 & 0.331 & 0.236 & 0.366 & 0.520 & 0.402 & 0.327 & 0.442 & 0.541 \\
&Robin\_S & 0.730& 0.614& 0.786& 0.993 & 0.192 & 0.109 & 0.208 & 0.368 & 0.325 & 0.231 & 0.360 & 0.512 & 0.436 & 0.365 & 0.471 & 0.571\\
&Robin\_D & 0.729& 0.612& 0.801& \cellcolor{cyan!25}\textbf{0.995} &  0.191 & 0.107 & 0.208 & 0.368 & 0.332 & 0.237 & \cellcolor{cyan!25}\textbf{0.370} & \cellcolor{cyan!25}\textbf{0.522} & 0.35 & 0.272 & 0.395 & 0.500\\
& Lion\_SD & 0.726& 0.605& 0.801& 0.993 & 0.193 & 0.110 & \cellcolor{cyan!25}\textbf{0.209} & \cellcolor{cyan!25}\textbf{0.369} & 0.330 & 0.235 & 0.366 & 0.519 & 0.433 & 0.363 & 0.471 & 0.562\\
&Lion\_SF & 0.725& 0.602& 0.801& 0.993 & \cellcolor{cyan!25}\textbf{0.194} & \cellcolor{cyan!25}\textbf{0.111} & \cellcolor{cyan!25}\textbf{0.209} & \cellcolor{cyan!25}\textbf{0.369} & 0.332 & 0.237 & 0.367 & 0.521 & 0.440 & 0.366 & 0.478 & 0.577 \\
& Tetra\_zero & 0.787 & 0.689 & 0.851 & 0.995 & 0.179 & 0.099 & 0.196 & 0.343 & 0.324 & 0.234 & 0.356 & 0.503 & 0.356 & 0.279 & 0.384 & 0.516 \\
&Tetra\_SF & 0.816 & 0.709 & 0.884 & \cellcolor{cyan!25}\textbf{0.995} & 0.181 & 0.102 & 0.195 & 0.345 & \cellcolor{cyan!25}\textbf{0.336} & \cellcolor{cyan!25}\textbf{0.245} & 0.368 & 0.518 & \cellcolor{cyan!25}\textbf{0.445} & \cellcolor{cyan!25}\textbf{0.367} & \cellcolor{cyan!25}\textbf{0.489} & \cellcolor{cyan!25}\textbf{0.593}  \\
&Tetra\_WSF & \cellcolor{cyan!25}\textbf{0.822} & \cellcolor{cyan!25}\textbf{0.731} & \cellcolor{cyan!25}\textbf{0.893} & 0.993 & 0.186 & 0.102 & 0.204 & 0.360 & 0.323 & 0.233 & 0.353 & 0.501 & 0.443 & 0.365 & 0.482 & 0.588 \\
\hline
\end{tabular}
\end{adjustbox}
\caption{Link prediction results for both low (D=32) and high (D=500) dimensional settings.  In each dimensional setting, \colorbox{cyan!25}{\textbf{numbers}} are the best results for each dataset.}
\label{tab:results for Nations, Diabetes, FB15k-237, YAGO-10}
\end{table*}

\begin{table*}[h!]
\centering
\resizebox{0.7\textwidth}{!}{ 
\begin{tabular}{|l|cccc|cccc|cccc|cccc|}
\hline
\multirow{2}{*}{\textbf{Model}} & \multicolumn{4}{c|}{\textbf{Nations}} & \multicolumn{4}{c|}{\textbf{Diabetes}} & \multicolumn{4}{c|}{\textbf{FB15k-237}} & \multicolumn{4}{c|}{\textbf{YAGO-10}}\\ \cline{2-17}
& MRR & H@1 & H@3 & H@10 & MRR & H@1 & H@3 & H@10 & MRR & H@1 & H@3 & H@10 & MRR & H@1 & H@3 & H@10 \\
\hline
KG-BERT & 0.592 & 0.420 & 0.716 & 0.982 & 0.063 & 0.022 & 0.057 & 0.143 & - & - & - & 0.420 & - & - & - & 0.292 \\
StAR & 0.545 & 0.348 & 0.677 & 0.950 & 0.102 & 0.033 & 0.103 & 0.229 & 0.296 & 0.205 & 0.322 & 0.482 & 0.254 & 0.169 & 0.271 & 0.426 \\
Our best & 0.822 & 0.731 & 0.893 & 0.995 & 0.194 & 0.111 & 0.209 & 0.369 & 0.336 & 0.245 & 0.370 & 0.522 & 0.445 & 0.367 & 0.489 & 0.593 \\
\hline
\end{tabular}
}
\caption{Comparison of LM-based baselines and our best results.  In each dimensional setting, \colorbox{cyan!25}{\textbf{numbers}} are the best results for each dataset.}
\label{tab:results for LM-based models}
\end{table*}

\paragraph{\textbf{Results of models purely based on pre-trained language models:}}
We also conduct a comparison of the results obtained by KG-BERT \cite{PLMyao2019kgbert} and StAR \cite{DBLP:conf/www/WangSLZW021} in Tables \ref{tab:results for LM-based models}.
Please note that these approaches do not explicitly provide embeddings for entities and relation labels, and thus the embedding dimension $D$ is not applicable.
We obtained these results by replicating their experiments on the other three datasets, while their results on FB15k-237 were obtained from the original papers.
The KG-BERT takes too much time running on the YAGO-10, so we can only report its intermediate testing result of Hits@10.
These results from both baselines are lower than our best results.
The reason for this is that approaches such as KG-BERT and StAR model triples as sequences of text using pre-trained language models, disregarding the structural information in KGs.
As a result, they encounter entity ambiguity problems on larger datasets. Unlike these approaches, KGE models that explicitly model entity embeddings naturally possess a unique representation for each entity.
In contrast, models like KG-BERT and StAR only have a unique representation for each word, and entities are represented as a sequence of words.
This become noisy with long entity descriptions, leading to noisy entity representations.

\paragraph{\textbf{The importance of integration in the hypercomplex space}}
To demonstrate the significance of integrating different features from texts in the hypercomplex space, we construct an ablation model, \textit{TransE\_Concat}, by concatenating four representations $\{W, F, D, S\}$ and treating it as the input for TransE.
The purpose of \textit{TransE\_Concat} is to assess whether the performance gain truly arises from the integration of the hypercomplex space or not.
We reused the low-dimensional results of TransE and the Dihedron version of our models from Tables \ref{tab:results for Nations, Diabetes, FB15k-237, YAGO-10}.
Simultaneously, we re-run the \textit{TransE\_Concat} model, and the results are presented in Table~\ref{tab:AblationResultsConcat}.
We observe that \textit{TransE\_Concat} outperforms TransE on the smallest dataset, Nations, but it falls short in comparison to our best model, Tetra\_WSF.
For the other larger datasets, \textit{TransE\_Concat} yield inferior results compared to Lion and even TransE.
These findings suggest that the simple concatenation strategy fails to efficiently capture the interaction between different features.
In contrast, our approach inherently accounts for such interactions due to the essence of the Dihedron product in Equation~\ref{dihedronproduct}, along with various geometric perspectives illustrated in Figure~\ref{fig: architecture}.

\begin{table*}[h!]
\centering
\begin{adjustbox}{width=\textwidth,center}
\begin{tabular}{|l|cccc|cccc|cccc|cccc|}
\hline
\multirow{2}{*}{\textbf{Model}} & \multicolumn{4}{c|}{\textbf{Nations}} & \multicolumn{4}{c|}{\textbf{Diabetes}} & \multicolumn{4}{c|}{\textbf{FB15k-237}} & \multicolumn{4}{c|}{\textbf{YAGO-10}} \\
\cline{2-17}
& MRR & H@1 & H@3 & H@10 & MRR & H@1 & H@3 & H@10 & MRR & H@1 & H@3 & H@10 &  MRR & H@1 & H@3 & H@10 \\
\hline
TransE & 0.684 & 0.542 & 0.779 & 0.990 & 0.166  & 0.089 & 0.182 & 0.322 & 0.274 & 0.197 & 0.298 & 0.428 & 0.368 & 0.284 & 0.403 & 0.534 \\
TransE\_Concat & 0.726& 0.612 & 0.786 & 0.983 & 0.161 & 0.088 & 0.174 & 0.312 & 0.257 & 0.182 & 0.278 & 0.406 & 0.149 & 0.087 & 0.160 & 0.272 \\
Lion\_SD & 0.736 & 0.624 & 0.801 & 0.988 & 0.167 & 0.090  & 0.180  & 0.330 & \cellcolor{cyan!25}\textbf{0.304} & \cellcolor{cyan!25}\textbf{0.217} & \cellcolor{cyan!25}\textbf{0.333} & \cellcolor{cyan!25}\textbf{0.478} & \cellcolor{cyan!25}\textbf{0.397} & \cellcolor{cyan!25}\textbf{0.314} & \cellcolor{cyan!25}\textbf{0.441} & \cellcolor{cyan!25}\textbf{0.554} \\
Lion\_SF & 0.727 & 0.605 & 0.801 & 0.993 & \cellcolor{cyan!25}\textbf{0.175} & \cellcolor{cyan!25}\textbf{0.097} & \cellcolor{cyan!25}\textbf{0.190} & \cellcolor{cyan!25}\textbf{0.340} & 0.301 & 0.214 & 0.330 & 0.475 & 0.395 & \cellcolor{cyan!25}\textbf{0.314} & 0.439 & 0.548 \\
Tetra\_WSF & \cellcolor{cyan!25}\textbf{0.780} & \cellcolor{cyan!25}\textbf{0.669} & \cellcolor{cyan!25}\textbf{0.858} & \cellcolor{cyan!25}\textbf{0.995} & 0.155 & 0.084 & 0.169 & 0.302 & 0.266 & 0.188 & 0.289 & 0.421 & 0.169 & 0.113 & 0.180 & 0.288 \\
\hline
\end{tabular}
\end{adjustbox}
\caption{Ablation study results of concatenating representations for the low (D=32) dimensional setting, where \colorbox{cyan!25}{\textbf{numbers}} are the best results for each dataset.}
\label{tab:AblationResultsConcat}
\end{table*}

\paragraph{\textbf{Comparison of Quaternion and Dihedron:}}
We conducted a comparison of the query representation using Quaternion and Dihedron on three variants of our model (Robin\_S, Lion\_SF, and Tetra\_WSF) by performing link prediction experiments on FB15k-237.
We reused the results of the Dihedron version of each model from Tables \ref{tab:results for Nations, Diabetes, FB15k-237, YAGO-10}, while rerunning the Quaternion version.
The results, as shown in Table \ref{tab:result of Quaternion and Dihedron}, indicate that Dihedron generally outperforms Quaternion, which is in line with our earlier claim in the Preliminary that Dihedron is a more expressive representation than Quaternion.

\begin{table}[h!]
\centering
\resizebox{0.7\textwidth}{!}{ 
    \begin{tabular}{|l|cccc|cccc|}
    \hline
    \multicolumn{1}{|c}{}&\multicolumn{4}{|c|}{\textbf{FB15k-237 (D=32)}} & \multicolumn{4}{|c|}{\textbf{FB15k-237 (D=500)}}\\
    \hline
    \multirow{1}{*}{\textbf{Model}} & MRR & H@1 & H@3 & H@10  &  MRR & H@1 & H@3 & H@10\\
    \hline
    Robin\_S (Quaternion) & 0.290 & 0.205 & 0.317 & 0.459 & 0.322 & 0.229 & 0.355 & 0.505 \\
    Robin\_S (Dihedron) & \cellcolor{cyan!25}\textbf{0.304} & \cellcolor{cyan!25}\textbf{0.222} & \cellcolor{cyan!25}\textbf{0.331} & \cellcolor{cyan!25}\textbf{0.465} & \cellcolor{cyan!25}\textbf{0.325} & \cellcolor{cyan!25}\textbf{0.231} & \cellcolor{cyan!25}\textbf{0.360} & \cellcolor{cyan!25}\textbf{0.512} \\
    \hline
    \hline
    Lion\_SF (Quaternion) & 0.295 & 0.208 & 0.322 & 0.465 & 0.320 & 0.228 & 0.353 & 0.505 \\
    Lion\_SF (Dihedron) & \cellcolor{cyan!25}\textbf{0.301} & \cellcolor{cyan!25}\textbf{0.214} & \cellcolor{cyan!25}\textbf{0.330} & \cellcolor{cyan!25}\textbf{0.475} & \cellcolor{cyan!25}\textbf{0.332} & \cellcolor{cyan!25}\textbf{0.237} & \cellcolor{cyan!25}\textbf{0.367} & \cellcolor{cyan!25}\textbf{0.521} \\
    \hline
    \hline
    Tetra\_WSF (Quaternion) & \cellcolor{cyan!25}\textbf{0.270} & \cellcolor{cyan!25}\textbf{0.192} & 0.294 & \cellcolor{cyan!25}\textbf{0.420} & 0.318 & 0.227 & 0.351 & 0.499 \\
    Tetra\_WSF (Dihedron) & 0.269 & \cellcolor{cyan!25}\textbf{0.192} & \cellcolor{cyan!25}\textbf{0.295} & 0.419 & \cellcolor{cyan!25}\textbf{0.328} & \cellcolor{cyan!25}\textbf{0.237} & \cellcolor{cyan!25}\textbf{0.361} & \cellcolor{cyan!25}\textbf{0.509} \\
    \hline
    \end{tabular}
}
\caption{Link prediction results on FB15k-237 with both dimensions. Each model use Quaternion product in Equation \ref{quatproduct} or Dihedron product in Equation \ref{dihedronproduct}. }
\label{tab:result of Quaternion and Dihedron}
\end{table}


\subsection{Effect Of Textual Information In Entity Representation}
\begin{figure}[h!]
\centering
\begin{minipage}[b]{0.45\linewidth}
\includegraphics[width=6cm,height=4.5cm]{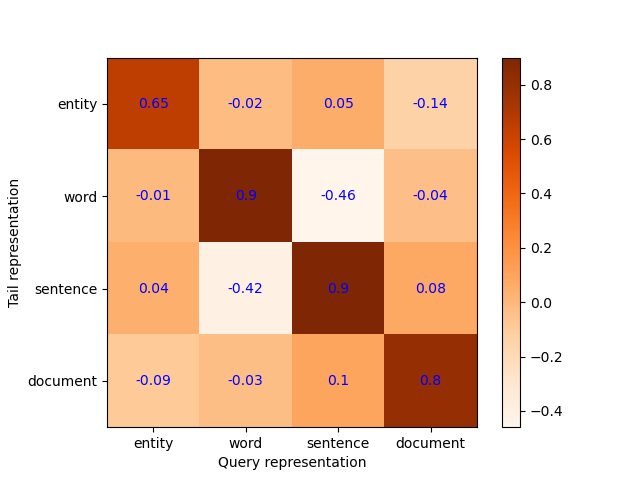}
\caption{Cosine similarities of the query and tail of triples on FB15k-237 .}
\label{fig:cossim_fb15k237}
\end{minipage}
\quad
\begin{minipage}[b]{0.45\linewidth}
\includegraphics[width=6cm,height=4.5cm]{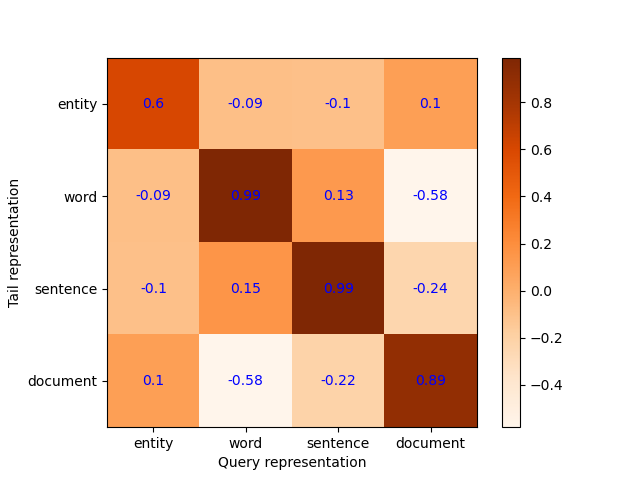}
\caption{Cosine similarities of the query and tail of triples on YAGO-10.}
\label{fig:cossim_yago10}
\end{minipage}
\quad
\end{figure}
In this visualization, we demonstrate the impact of pre-trained language models on the Tetra\_WSF model's performance.
We selected two datasets, FB15k-237 and YAGO-10, and computed the average cosine similarities on all testing triples of each dataset.
The heat maps in Figure~\ref{fig:cossim_fb15k237} and \ref{fig:cossim_yago10} show the cosine similarities between the four parts of our 4*D dimensional representation: \textit{entity embedding}, \textit{word embedding}, \textit{sentence embedding}, and \textit{document embedding}.
By examining the similarities on the diagonal of the heat maps, we can assess the contribution of each part when matching the query and the tail in the link prediction task.

Among the four types of embeddings, the similarities between word-word, sentence-sentence, and document-document embeddings are higher than the similarity between entity-entity embeddings, indicating that semantic information at the word, sentence, and document levels are helpful in matching queries and tails.
We conclude that all three levels of textual information from entity descriptions capture important semantic information during the matching between queries and tails.

\subsection{Contribution of individual sentences}
In this analysis, we aim to investigate how our Robin\_S model leverages sentence-level representations by selecting important sentences. Specifically, we evaluate the model's performance on the YAGO-10 dataset for link prediction and collect sentences from the descriptions of the head and tail entities for each triple.
We then compute the importance of each sentence using the Shapley value method \cite{shapley201617} and rank them in descending order of importance. 
Finally, we manually inspect the semantics of each highly ranked sentence, and conclude that our model effectively exploits sentence-level information for the corresponding triple if a semantically important sentence receives a high rank.

Table~\ref{tab:ablation sentence shapley} presents the top-3 important sentences from the entity descriptions, as identified by the Shapley value, in the YAGO-10 dataset.
For the first triple (MarsCallahan, created, Zigs(film)), the top-ranked sentence comes from the tail description and contains the keywords ``directed by Mars Callahan,'' which is about the head entity.
For the second triple (MargaretOfGeneva, isMarriedTo, ThomasCountOfSavoy), the top-1 and top-2 sentences refer to the definition of both entities.
Notably, the keywords ``escorting'' and ``carried her off'' in the top-3 sentence provide information about the relation label ``isMarriedTo.''
These examples demonstrate how our model effectively exploits sentence-level information from the entity descriptions for link prediction.

\begin{table*}[h!]
\centering
\begin{adjustbox}{width=\textwidth,center}
\begin{tabular}{|c|c|c|c|}
\hline
Triple & \makecell{Sentence \\ rank} & \makecell{Sentence\\ source} & Sentence \\
\hline
\multirow{3}{*}{\makecell[l]{MarsCallahan,\\ created,\\ Zigs(film)}} & 1 & tail & \makecell{\textbf{Zigs} is a 2001 English language drama starring Jason Priestley Peter Dobson and Richard Portnow \\and \textbf{directed by Mars Callahan}.}\\
        & 2 & tail & The film received an r rating by the MPAA.\\
        & 3 & head & At the age of eleven Callahan toured with a children's musical group through thirty-seven states.\\
\hline
\hline
\multirow{3}{*}{\makecell[l]{MargaretOfGeneva,\\ isMarriedTo,\\ ThomasCountOfSavoy}} & 1 & tail & \makecell{\textbf{Thomas Tommaso I was count of savoy} from 1189 to 1233 he is sometimes numbered Thomas I \\to distinguish him from his son of the same name who governed savoy but was not count.}\\
& 2 & head & \textbf{Margaret of Geneva} 1180-1252 countess of savoy was the daughter of William I count of Geneva.\\
& 3 & head & When her father was \textbf{escorting} her to France in May 1195 Thomas I of savoy \textbf{carried her off}. \\
\hline

\end{tabular}
\end{adjustbox}
\caption{Examples about sentence contribution of our model, where a sentence with higher rank (smaller number) is more important. The source of the sentence indicates whether a sentence comes from the description of a head or tail entity. }
\label{tab:ablation sentence shapley}
\end{table*}

\section{Conclusion}
In this study, we investigated the effectiveness of incorporating multi-level textual information by using multiple pre-trained language models in a KGE model.
Our novel KGE model based on the Dihedron algebra captures the interactions between embeddings from pre-trained language models, resulting in better representations for incomplete KGs.
Our experiments demonstrate that the incorporation of more pre-trained language models is beneficial for extracting information from text in KGs, particularly on small or sparse KGs.
While other recent work \cite{DBLP:conf/www/LiZXZX23} focuses on unifying information from multiple sources, our work is the first to explore the use of multiple pre-trained language models in the representation learning of KGs.
In future work, we plan to investigate the incorporation of multi-source information in multi-hop KG completion scenarios.
Beside the KG completion task, we also plan to apply our model to various other tasks.
For instance, we can adapt it to predict entity types given a schema or to facilitate complex logical query answering.

\paragraph*{Supplemental Material Statement:} The datasets and our implementation are available at \url{https://github.com/ZihaoWang/text_enhanced_KGE}.
The Appendix can be found in our full Arxiv version \url{https://arxiv.org/abs/2208.02743}.

\section*{Acknowledgement}
The authors thank the International Max Planck Research School for Intelligent Systems (IMPRS-IS) for supporting Zihao Wang. 
Zihao Wang and Mojtaba Nayyeri have been funded by the German Federal Ministry for Economic Affairs and Climate Action under Grant Agreement Number 01MK20008F (Service-Meister) and ATLAS project funded by Bundesministerium für Bildung und Forschung (BMBF). 

\appendix
\newpage
\appendix

\section{Details of Datasets}
\label{Appendix:DatasetsDetails}
The \textbf{NATIONS} dataset \cite{kok2007statistical} is a small-scale KG that consists of relation labels between countries.
In this KG, the entity names are provided, and we manually collect entity descriptions from Wikipedia.
The \textbf{Diabetes} dataset is a medical-domain KG \cite{10.1093/bioinformatics/btac085}, which contains both names and descriptions for entities.
The \textbf{FB15k-237} \cite{DBLP:conf/acl-cvsc/ToutanovaC15} is a subset of the Freebase dataset FB15k \cite{bordes2013translating} with data leakage caused by inverse relations resolved.
We collect triples and entity descriptions from \cite{xie2016DKRL} and entity names from the dataset\footnote{\label{fbname}https://git.uwaterloo.ca/jimmylin/BuboQA-data/raw/master/FB5M.name.txt.bz2}.
Lastly, the \textbf{YAGO-10} is a subset of the YAGO dataset \cite{DBLP:journals/corr/abs-1809-01341}. We collect triples and entity names from the repository\footnote{https://github.com/TimDettmers/ConvE} and entity descriptions from the repository\footnote{https://github.com/pouyapez/mkbe}.

\section{Environment and Hyperparameters}
\label{Appendix:Hyperparameters}
Our models are implemented using PyTorch \cite{DBLP:conf/nips/PaszkeGMLBCKLGA19} and built on top of Chami's framework \cite{DBLP:conf/acl/ChamiWJSRR20}.
We utilize several pre-trained language models, namely \textit{Word2Vec} \cite{mikolov2013distributed} and \textit{Doc2Vec} \cite{DBLP:journals/corr/LeM14} from the Python Gensim library, \textit{fastText}\cite{bojanowski2017enriching}\footnote{https://dl.fbaipublicfiles.com/fasttext/vectors-wiki/wiki.simple.zip}, and \textit{Sentence Transformer} with the pre-trained model \textit{distilbert-base-nli-mean-tokens}\footnote{https://huggingface.co/sentence-transformers}.

The hyperparameters we use for both low($D=32$) and high dimension($D=500$) setup are shown in Table~\ref{tab:hyperparam}, where $b$ is the batch size, $l_r$ is the learning rate and $n$ is the negative sampling size.
Specifically, $n=-1$ indicates that we use full cross-entropy loss that takes all other triples in the dataset as negative triples instead of performing negative sampling.
\begin{table}[h]
\centering
\begin{tabular}{|l|c|c|c|c|} 
 \hline
Dataset & D & b & $l_r$ & n \\
 \hline
 \multirow{2}{*}{NATIONS}&32 & 400 & 0.01 & 100 \\ 
 \cline{2-5}
 &500 & 400 & 0.01 & 10 \\
 \hline
\multirow{2}{*}{Diabetes}& 32 & 100 & 0.25 & -1 \\ 
 \cline{2-5}
 &500 & 100 & 0.05 & 100 \\ 
 \hline
\multirow{2}{*}{FB15k-237}&32 & 100 & 0.01 & 100 \\ 
 \cline{2-5}
 &500 & 100 & 0.01 & 100 \\ 
 \hline
 \multirow{2}{*}{YAGO-10}&32 & 400 & 0.25 & 100 \\ 
 \cline{2-5}
 &500 & 100 & 0.01 & -1 \\\hline
\end{tabular}
\caption{Optimal hyperparameters for all models on each dataset.}
\label{tab:hyperparam}
\end{table}

\section{Semantic Clustering Of Entity Embedding}
\begin{figure}[h!]
    \centering
    \includegraphics[width=12cm,height=6cm]{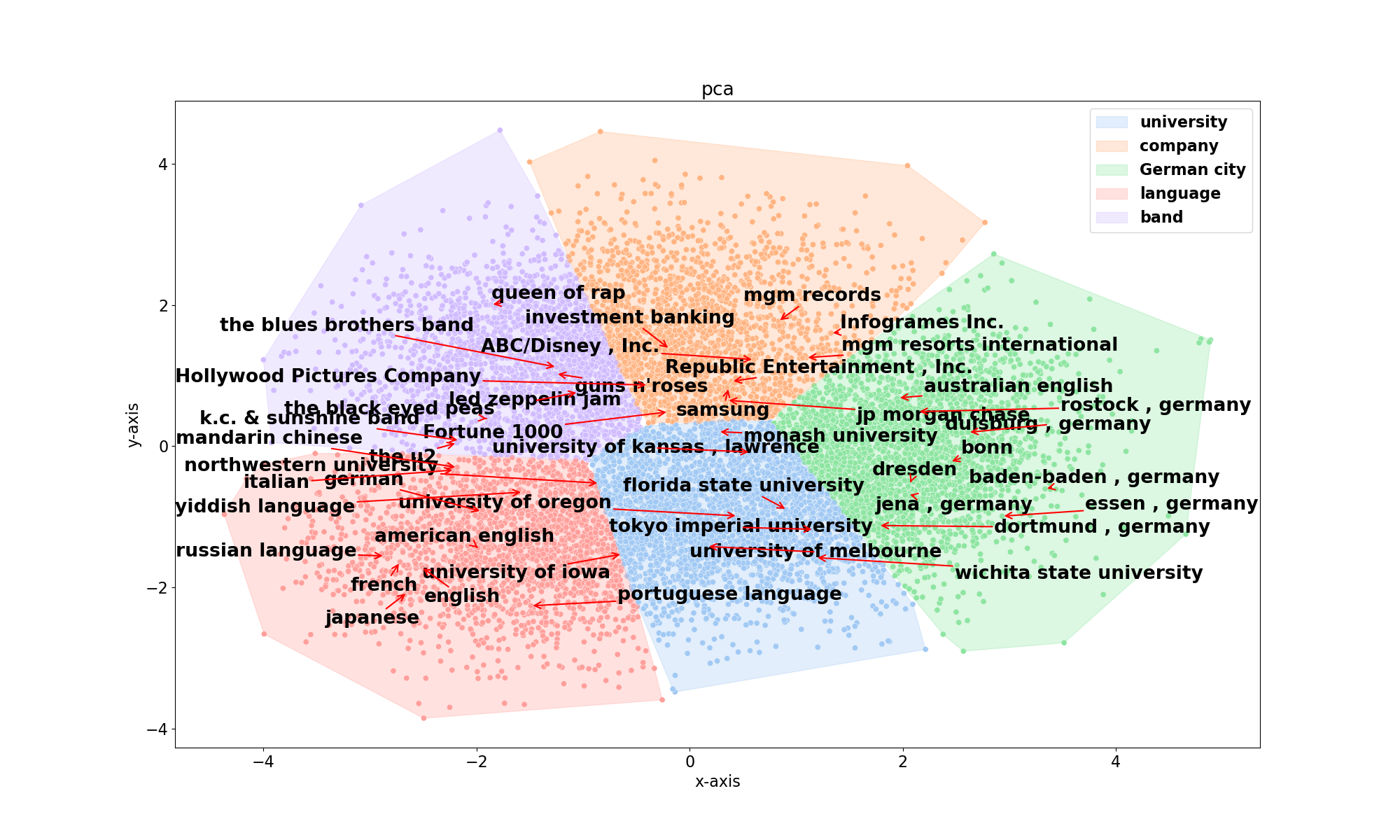}
    \caption{The clustering of trained entity embeddings from our model Robin\_S model on the FB15k-237 dataset where dimension $D=32$. Each cluster in the figure represents a particular topic, each point corresponds to an entity in the dataset, and each red arrow points to the corresponding point of an entity name.}
    \label{fig:clustering}
\end{figure}

We present an analysis of the semantics of learned entity embeddings from our best-performing model, Robin\_S, on the FB15k-237 dataset.
The goal is to determine if semantically similar entities also have similar representations in the learned embedding space.
Specifically, we apply K-means clustering and t-SNE \cite{van2008visualizing} dimension reduction to the learned entity embeddings, resulting in a 2-dimensional visualization shown in Figure~\ref{fig:clustering}.
The results of the analysis show that semantically related entities are indeed clustered together in the learned embedding space.
For instance, universities and languages are separated into distinct blue and red clusters, respectively.
This suggests that our model is capable of capturing meaningful semantic relationships between entities in the KG.

%
%
%
\newpage
\bibliographystyle{splncs04}
\bibliography{ref.bib}

\end{document}